\documentclass{article}

\usepackage{microtype}
\usepackage{graphicx}
\usepackage{booktabs} 

\usepackage{hyperref}

\usepackage[accepted]{icml2020}

\icmltitlerunning{Integrating Logical Rules Into Neural Multi-Hop Reasoning for Drug Repurposing}

\usepackage{graphicx}
\usepackage{subcaption}
\usepackage[export]{adjustbox}
\usepackage{amsmath, amssymb}
\usepackage{dsfont}
\DeclareMathOperator*{\argmax}{arg\,max}

\begin{document}

\twocolumn[
\icmltitle{Integrating Logical Rules Into Neural Multi-Hop Reasoning\\ for Drug Repurposing}



\icmlsetsymbol{equal}{*}

\begin{icmlauthorlist}
\icmlauthor{Yushan Liu}{equal,siemens,lmu}
\icmlauthor{Marcel Hildebrandt}{equal,siemens,lmu}
\icmlauthor{Mitchell Joblin}{siemens}
\icmlauthor{Martin Ringsquandl}{siemens}
\icmlauthor{Volker Tresp}{siemens,lmu}
\end{icmlauthorlist}

\icmlaffiliation{siemens}{Siemens AG, Corporate Technology, Munich, Germany}
\icmlaffiliation{lmu}{Ludwig Maximilian University of Munich, Munich, Germany}

\icmlcorrespondingauthor{Yushan Liu}{yushan.liu@siemens.com}

\icmlkeywords{Multi-hop reasoning, Reinforcement learning, Machine learning with background knowledge, Logical rules, Biomedical knowledge graphs}

\vskip 0.3in
]



\printAffiliationsAndNotice{\icmlEqualContribution} 


\begin{abstract}




The graph structure of biomedical data 
differs from those in typical knowledge graph benchmark tasks. A particular property of biomedical data is the presence of long-range dependencies, which can be captured by patterns described as logical rules. 
We propose a novel method that combines these rules with a neural multi-hop reasoning approach that uses  reinforcement learning. 
We conduct an empirical study based on the real-world task of drug repurposing by formulating this task as a link prediction problem.
We apply our method to the biomedical knowledge graph Hetionet and show that our approach outperforms several baseline methods.


\end{abstract}

\section{Introduction}
\label{sec:introduction}
\if false
\begin{itemize}
    \item KGs in the biomedical domain; long range dependencies, specific characteristics
    \item drug repurposing (Remdesivir?); time + cost savings
    \item embedding-based methods vs path-based methods vs rules; specific characteristics of biomedical KGs
    \item explainability?
    \item domain knowledge and structured knowledge representation
    \item importance of Hetionet; connection to linked open data? 
\end{itemize}
\fi
Advancements in low-cost high-throughput sequencing and data acquisition technologies have given rise to a massive proliferation of data describing biological systems. 
Biomedical knowledge graphs (KGs) are becoming increasingly popular as backbones for artificial intelligence tasks such as personalized medicine, predictive diagnosis, and drug discovery \cite{Doerpinghaus2019}. 

\begin{figure}[t!]
    \centering
    \includegraphics[width=0.8\columnwidth]{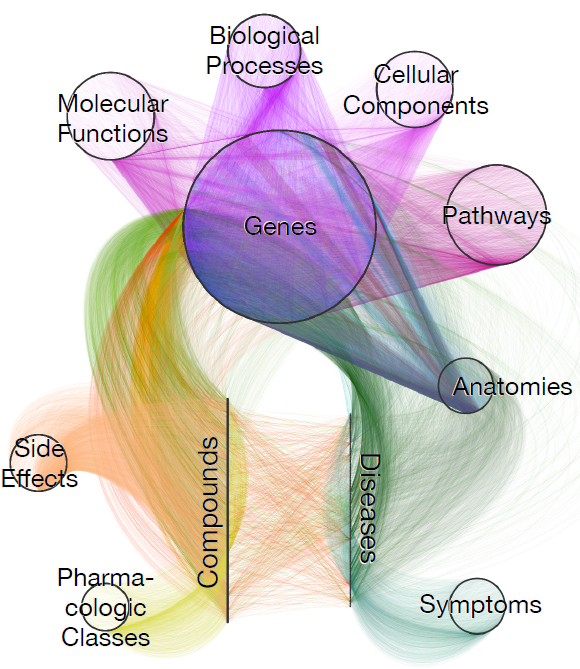}
    \caption[dummy]{
    Visualization of the heterogeneous biomedical network Hetionet\footnotemark.
    }
    \label{fig:biomedical_network}
\end{figure}
\footnotetext{\copyright\;Himmelstein et al.~\yrcite{himmelstein2017systematic}, licensed under \href{https://creativecommons.org/licenses/by/4.0/}{CC BY 4.0}.}
From a machine learning perspective, reasoning on biomedical KGs presents new challenges for existing approaches because of the unique structural characteristics of the graphs. One challenge arises due to the highly coupled nature of entities in biological systems that leads to many high-degree and densely interlinked entities.
A second challenge is the requirement of information beyond second-order neighborhoods for reasoning about the relationship between two entities~\cite{himmelstein2017systematic} so that 
approaches where long-range interactions are incorporated only via node embeddings (e.\,g., RESCAL~\cite{nickel2011three}, TransE~\cite{bordes2013translating}) tend to underperform. 
Unfortunately, approaches that explicitly take the entire multi-hop neighborhoods into account (e.\,g., graph convolutional models, R-GCN~\cite{schlichtkrull2018modeling}), often have diminishing performance beyond two-hop neighborhoods (i.\,e., more than two convolutional layers). Furthermore,  high-degree entities can cause the aggregation operations to smooth out the signals. Alternatively, symbolic reasoning approaches (e.\,g., RuleN~\cite{meilicke2018fine}, AnyBURL~\cite{meilicke2019anytime}) learn logical rules and employ them during inference. However, due to the massive scale and diverse topologies of many real-world KGs, combinatorial complexity often prevents the usage of symbolic approaches. Also, logical inference has difficulties handling noise in the data. Recently, path-based reasoning methods have become popular, and they present a seemingly ideal balance for combining information over multi-hop neighborhoods. 

We propose a novel neuro-symbolic KG reasoning approach that combines path-based approaches with representation learning and logical rules.
These rules can be either mined from data or obtained from domain experts. Inspired by existing methods~\cite{minerva,lin2018multi,hildebr2020scene, hildebrandt2020reasoning}, we use reinforcement learning to train an agent to conduct policy-guided random walks on a KG. We propose a modification by introducing a reward function that allows the agent to leverage background knowledge formalized as metapaths. 
In summary, our paper makes the following contributions:
\begin{itemize}
    \item We propose a novel neuro-symbolic approach that combines neural multi-hop reasoning based on reinforcement learning with logical rules.
    \item We conduct an empirical study of several state-of-the-art algorithms applied to a large biomedical KG.
    \item We show that our proposed approach outperforms state-of-the-art alternatives on a highly relevant biomedical prediction task (drug repurposing).
\end{itemize}
\noindent

As an application of our method, we focus on the drug repurposing problem, which is characterized by finding new treatment targets for existing drugs. By repurposing existing drugs, available knowledge about drug-disease-interactions can be leveraged to reduce time and cost for developing new drugs significantly. A recent example is the repositioning of the medication remdesivir for the novel coronavirus disease COVID\nobreakdash-19. 
We aim at generating candidates for the drug repurposing task with machine learning reasoning methods and  formulate the task as a link prediction problem, where both compounds and diseases correspond to entities in a KG. 

\section{Notation}
\label{sec:background}
\begin{figure}
    \centering
    \includegraphics[width=0.9\columnwidth]{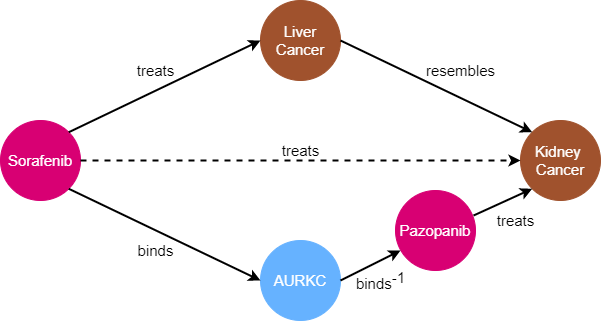}
    \caption{Subgraph of Hetionet that illustrates the drug repurposing use case: The two paths that connect the chemical compound sorafenib and the disease kidney cancer can be used to predict a direct edge between the two entities.}
    \label{fig:kg_example}
\end{figure}
Let $\mathcal{E}$ denote the set of entities in a KG and $\mathcal{R}$ the set of binary relations. Elements in $\mathcal{E}$ correspond to biomedical entities including, e.\,g., chemical compounds, diseases, and genes. Each entity belongs to a unique type in $\mathcal{T}$, defined by the mapping $\tau : \mathcal{E} \rightarrow \mathcal{T}$. For example,  $\tau(\textit{AURKC}) =~\textit{Gene}$ indicates that the entity \textit{AURKC} has type \textit{Gene}. 
We define a KG $\mathcal{KG} \subset \mathcal{E} \times \mathcal{R} \times \mathcal{E} $ as a collection of triples of the form $(h, r, t)$, which consists of head, relation, and tail. 
Head and tail entities correspond to nodes in the graph, while the relation indicates the type of edge between them. For any relation $r \in \mathcal{R}$, we denote the corresponding inverse relation with $r^{-1}$  (i.\,e., $(h, r, t)$ is equivalent to $(t, r^{-1}, h)$). Triples in $\mathcal{KG}$ are interpreted as true known facts. For example, the triple 
$(\textit{Sorafenib}, \textit{treats}, \textit{Liver Cancer}) \in \mathcal{KG}$ 
in Figure \ref{fig:kg_example} corresponds to the fact that the kinase inhibitor drug sorafenib is approved for the treatment of liver cancer. 

We further distinguish between two types of paths: instance paths and metapaths.
An instance path of length $T \in \mathbb{N}$ on $\mathcal{KG}$ is given by a sequence  
$$(e_1 \xrightarrow{r_1} e_2 \xrightarrow{r_2} \cdots \xrightarrow{r_{T}} e_{T+1}),$$ 
where $(e_i, r_i, e_{i+1}) \in \mathcal{KG}$. Moreover, we call 
$$(\tau(e_1) \xrightarrow{r_1} \tau(e_2) \xrightarrow{r_2} \cdots \xrightarrow{r_{T}} \tau(e_{T+1}))$$ a metapath. 
For example,
$$(\textit{Sorafenib} \xrightarrow{\textit{treats}} \textit{Liver Cancer} \xrightarrow{\textit{resembles}} \textit{Kidney Cancer})$$ constitutes an instance path of length 2, where $$(\textit{Compound} \xrightarrow{\textit{treats}} \textit{Disease} \xrightarrow{\textit{resembles}} \textit{Disease})$$ is the corresponding metapath. 

Logical rules (e.g., the commonly used Horn clauses) 
are usually written in the form $\textit{head} \leftarrow \textit{body}$. 
The head can be written out as a triple, and the body can be expressed as a metapath. Define $\textit{CtD} :=~(\textit{Compound}, \textit{treats}, \textit{Disease})$.
Then, a rule with respect to edges of type \textit{treats} is of the generic form
$$ \textit{CtD}  \leftarrow \left(\textit{Compound} \xrightarrow{r_1} \textit{Type}_2 \xrightarrow{r_2} \dots \xrightarrow{r_T} \textit{Disease} \right).
$$
In particular, the body of a rule corresponds to a metapath starting at a compound and terminating at a disease. The goal is to find instance paths where the corresponding metapaths match the body of a rule to predict a new relation between the source and the target of the instance path. The confidence of a rule indicates how often a rule is correct and is defined as the rule support divided by the body support in the data. 

\if false
For example, consider the rule
\begin{align*}
\textit{CtD} &\leftarrow \\ &(\textit{Compound} \xrightarrow{\textit{binds}} \textit{Gene} \xrightarrow{\textit{binds}^{-1}} \\& \textit{Compound} \xrightarrow{\textit{treats}} \textit{Disease}).
\end{align*}
The metapath of the following instance path
\begin{align*}
&(\textit{Sorafenib} \xrightarrow{\textit{binds}} \textit{AURKC} \xrightarrow{\textit{binds}^{-1}} \textit{Pazopanib} \\ &\xrightarrow{\textit{treats}} \textit{kidney cancer})
\end{align*}
matches the rule body, suggesting that sorafenib can also treat kidney cancer. 
\fi

\if false
\subsection{Related Work}
Even though real-world KGs contain a massive number of triples, they are still expected to suffer from incompleteness in the sense that true facts are missing. Therefore, link prediction (also known as KG completion) is a common reasoning task on KGs. Moreover, many classical artificial intelligence tasks such as  recommendation problems or question answering can be rephrased in terms of link prediction in KGs. In this work, we frame the drug repurposing task as a link prediction problem on a biomedical KG. 

Symbolic approaches have a far-reaching tradition in the context of knowledge acquisition and reasoning. The task of reasoning with logical rules has been addressed in areas such as Markov logic networks \cite{richardson2006markov} or inductive logic programming \cite{muggleton1991inductive}. However, such techniques typically do not scale well to modern, large-scale KGs. Recently, novel methods such as RuleN \cite{meilicke2018fine} and its successor AnyBURL \cite{meilicke2019anytime} have been proposed that achieve state-of-the-art performance on popular benchmark datasets such as FB15k-237 \cite{toutanova2015representing} and WN18RR \cite{dettmers2018convolutional}. 

Subsymbolic approaches map symbolic entities such as nodes and edges in KGs to low-dimensional vector representations known as embeddings. Then, the likelihood of missing triples is approximated by a classifier that operates on the embedding space. Popular embedding-based link prediction methods include translational methods such as TransE \cite{bordes2013translating} and TransR \cite{lin2015learning} or the tensor factorization approaches RESCAL \cite{nickel2011three} and ComplEx \cite{trouillon2016complex}. Moreover, R-GCNs~\cite{schlichtkrull2018modeling} were proposed, which extend graph convolutional networks \cite{kipf2016semi} to multi-relational graphs.

Despite achieving good results on the link prediction task, a fundamental problem of many embedding-based methods is their non-transparent nature in the sense that it remains hidden to the user what contributed to the  predictions. Moreover, most embedding-based methods cannot capture the compositionality expressed by long reasoning chains. This often limits their applicability to complex reasoning tasks. Multi-hop reasoning methods (also known as path-based methods) follow a different philosophy than embedding-based methods. The underlying idea is to infer missing knowledge based on extracting paths from the KG. Thereby, these methods come with an inherent transparency mechanism by providing explicit reasoning chains that can be analyzed by the user. The pioneering method Path Ranking Algorithm (PRA) \cite{lao2010relational} formulates the link prediction task as a maximum likelihood classification based on paths sampled from a nearest neighbor random walk. Xiong et al.\;extend the idea from PRA and frame the task of path extraction as a reinforcement learning problem~\cite{wenhan_emnlp2017}. They propose DeepPath, which substitutes the nearest neighbor random walk from PRA with a policy-guided random walk. The method that we propose in this work is an extension of the path-based method MINERVA \cite{minerva}. The underlying mechanism is to train a reinforcement learning agent to perform a policy-guided random walk until the answer entity to a query is reached. The goal is that the agent's policy implicitly encodes fuzzy logical rules that generalize to unseen test instances.

One of the drawbacks of existing policy-guided walk methods is that the agent might receive noisy reward signals based on corrupted or spurious triples that lead to the correct answer entity during training, but lower the generalization capabilities during testing. Moreover, biomedical KGs often exhibit both long-range dependency structures and high-degree hub nodes (see Section~\ref{sec:dataset}). 
The combination of these two properties makes it difficult for MINERVA's agent to navigate over biomedical KGs and extend a path in the most promising way. As a remedy, we propose the incorporation of known, effective logical rules via a novel reward function. This can help to denoise the reward signal and guide the agent on long paths with high-degree nodes.


\fi

\section{Our Method}
\label{sec:our_method}

We pose the task of drug repurposing as a link prediction problem based on graph traversal. Starting at a query entity (e.g., a compound to be repurposed), an agent performs a walk on the graph by sequentially transitioning to a neighboring node. The decision of which transition to make is determined by a stochastic policy. Each subsequent transition is added to the current path, extending the reasoning chain, until a finite number of transitions is reached. 
The general approach is inspired by the reinforcement learning method MINERVA~\cite{minerva}, with our primary contribution coming from the incorporation of logical rules into the training process.

\sloppy 
The state of the environment consists of the entity $e_t$ where the agent is located at time $t$, the source entity $e_c$, and the target entity $e_d$, where $e_c$ and $e_d$ correspond to the compound that we aim to repurpose and the target disease, respectively. 
Thus, a state $S_t$ for time $t \in \mathbb{N}$ is represented by $S_t := \left(e_t, e_c, e_d\right)$. The agent is given no information about the target disease so that the observed part of the state space is given by $\left(e_t, e_c\right) \in \mathcal{E}^2$.
Let $\boldsymbol{e} \in \mathbb{R}^{d}$ denote the embedding of entity $e$ and $\boldsymbol{r} \in \mathbb{R}^{d}$ the embedding of relation $r$.
The set of available actions contains all outgoing edges from the node $e_t$ with the corresponding target nodes and the option to stay at the current node with no transition. We denote with $A_t \in \mathcal{A}_{S_t}$ the action that the agent performed at time $t$. The environment evolves deterministically by updating the state according to the previous action. 

 The agent encodes previous actions via a multi-layered LSTM \cite{hochreiter1997long}
\begin{equation}
\label{eq:lstm_agent}
        \boldsymbol{h}_t = \text{LSTM}\left(\left[\boldsymbol{a}_{t-1}, \boldsymbol{e}_c\right]\right),
\end{equation}
where $\boldsymbol{a}_{t-1} := \left[\boldsymbol{r}_{t-1},\boldsymbol{e}_{t}\right] \in \mathbb{R}^{2d}$ corresponds to the vector space embedding of the previous action (or the zero vector at time $t=0$).
The action distribution is given by
\begin{equation}
\label{eq:policy_agent}
\boldsymbol{d}_t = \text{softmax}\left(\boldsymbol{A}_t \left(\boldsymbol{W}_2\text{ReLU}\left(\boldsymbol{W}_1 \boldsymbol{h}_t\right)\right)\right),
\end{equation}
where $\boldsymbol{W_1}$ and $\boldsymbol{W_2}$ are weight matrices and the rows of $\boldsymbol{A}_t \in \mathbb{R}^{\vert \mathcal{A}_{S_t} \vert \times 2d}$ contain the latent representations of all admissible actions from $S_t$. An action $A_t \in \mathcal{A}_{S_t}$ is sampled according to
$
    A_t \sim \text{Categorical}\left(\boldsymbol{d}_t\right).
$
Overall, $T$ transitions are sampled, resulting in a path denoted by $$
    P := (e_c \xrightarrow{r_1} e_2 \xrightarrow{r_2} \dots \xrightarrow{r_{T}} e_{T+1}) ,
$$ 
where $T$ is the maximum path length. 
Equations \eqref{eq:lstm_agent} and \eqref{eq:policy_agent} induce a stochastic policy, represented by $\pi_{\boldsymbol{\theta}}$ where $\boldsymbol{\theta}$ denotes the set of all trainable parameters, including all entity and relation embeddings.

Furthermore, let $\mathcal{M}~=~\{M_1, M_2, \dots, M_m\}$ be the set of metapaths, where each element corresponds to the body of a rule. 
For every metapath $M$, we assign a score $S(M)\in \mathbb{R}$ that indicates a quality measure of the corresponding rule, such as the confidence or the support with respect to making a correct prediction. For a path $P$, we denote with $\Tilde{P}$ the corresponding metapath. 

During training, a terminal reward is computed according to
\begin{equation*}
\label{eq:rewards}
    R =  \mathbb{I}_{\{e_{T+1} = e_d\}}\left(1 + \lambda \sum_{i = 1}^m S(M_i) \mathbb{I}_{\{\Tilde{P} = M_i\}}\right).
\end{equation*}
The first term indicates whether the agent has reached the correct target disease. The second term checks whether the metapath corresponds to the body of a rule and adds to the score accordingly. Heuristically speaking, we want to reward the agent with a higher score for extracting a metapath that corresponds to a body. The hyperparameter $\lambda \geq 0$ balances the two components of the reward. For $\lambda = 0$, we recover MINERVA.

We employ REINFORCE \cite{williams1992simple} to maximize the expected rewards. Thus, the agent's maximization problem is given by
\begin{equation}
\label{eq:objective_agent}
    \argmax_{\boldsymbol{\theta}} \mathbb{E}_{e_c \sim \mathcal{E}_c}\mathbb{E}_{A_1, A_2, \dots, A_{T} \sim \pi_{\boldsymbol{\theta}}}\left[R \mid e_c \right]  ,
\end{equation}
where $\mathcal{E}_c$ denotes the true underlying distribution of the set of chemical compounds.

\section{Experiments}
\label{sec:experiments}
\subsection{Dataset}
\label{sec:dataset}
Hetionet ~\cite{himmelstein2017systematic} is a biomedical KG that integrates data from 29 highly reputable and cited public databases.
It consists of 47,031 entities with 11 different types and 2,250,197 edges with 24 different types. 
We aim to predict edges with type \textit{treats} between entities that correspond to compounds and diseases. The goal is to perform candidate ranking according to the likelihood of successful drug repurposing in a novel treatment application.  
There are 1552 compounds and 137 diseases in Hetionet with 775 observed links of type \textit{treats} between compounds and diseases. 

\if false
Figure~\ref{fig:metagraph} illustrates the schema and shows the different types of entities and possible relations between them.

\begin{figure}[t!]
    \centering
    \includegraphics[width=0.75\textwidth]{figures/metagraph.png}
    \caption{Metagraph of Hetionet \cite{himmelstein2017systematic}\textsuperscript{1}: Hetionet has 11 different entity types and 24 possible relations between them.}
    \label{fig:metagraph}
    \vspace{2mm}
\end{figure}

Hetionet differs in many aspects from the standard benchmark datasets that are typically used in the KG reasoning literature. Table \ref{tab:datasets} summarizes the basic statistics of Hetionet along with the popular benchmark dataset FB15k-237 \cite{toutanova2015representing} and WN18RR \cite{dettmers2018convolutional}. One of the major differences between Hetionet and the two other benchmark datasets is the density of  triples with respect to the number of entities. It means that the average node degree in Hetionet is significantly higher than in the other two KGs. In addition, Figure  \ref{subfig:degrees} illustrates the average degrees for the different entity types in Hetionet. It is apparent that entities of type \textit{Anatomy} are densely connected hub nodes. In addition, entities of type \textit{Gene} have an average degree of around 123. This plays a crucial role for our application since many relevant paths that connect \textit{Compound} and \textit{Disease} traverse entities of type \textit{Gene} (see Figure \ref{fig:metagraph} and Table \ref{tab:metapaths}). We will discuss in Section~\ref{subsec:discussion} further how particularities of Hetionet impose challenges for existing KG reasoning methods.

\begin{table}[t!]
\caption{Comparison of Hetionet with the two benchmark datasets FB15k-237 and WN18RR.}
\center
\resizebox{\columnwidth}{!}{
\begin{tabular}{ c  c  c  c c}
 \hline
 Dataset & Entities & Relations & Triples & Avg. degree \\
 \hline \hline
Hetionet & 47,031 & 24 & 2,250,197 & 95.8\\
FB15k-237 & 14,541 & 237 & 310,116 & 19.7\\
WN18RR & 40,943 & 11 & 93,003 & 2.2 \\
 \vspace{0.2cm}
\end{tabular}
}
\label{tab:datasets}
\end{table}
\if false
\begin{figure}[t]
    \begin{subfigure}{0.45\textwidth}
        \includegraphics[width=\linewidth]{figures/entity_counts.png} 
        \caption{Entity counts}
        \label{subfig:counts}
    \end{subfigure}
    \begin{subfigure}{0.45\textwidth}
        \includegraphics[width=\linewidth,center]{figures/entity_degree.png}
        \caption{Average node degrees}
        \label{subfig:degrees}
    \end{subfigure}
    \caption{Comparison of the total counts (a) and the average node degrees (b) according to each entity type in Hetionet.}
    \label{fig:counts_degrees}
\end{figure}
\fi
\fi


\subsection{Metapaths as Background Information}

Himmelstein et al.~\yrcite{himmelstein2017systematic} compiled a list of 1206 metapaths corresponding to various pharmacological efficacy mechanisms that connect entities of type \textit{Compound} with entities of type \textit{Disease}. Through hypothesis testing and domain expertise, they identified $31$ effective metapaths that served as features for a logistic regression model. 
Out of these metapaths, we select the 10 metapaths as background information that have at most path length 3 and exhibit positive regression coefficients, indicating their importance for predicting drug efficacy. The metapaths are included as rule bodies in $\mathcal{M}$, where the rule head is always \mbox{(\textit{Compound}, \textit{treats}, \textit{Disease})}. We estimate the confidence score for each rule by sampling 10,000 paths whose metapaths correspond to the rule body and use the confidence for the score $S(M)$  (see Section {\ref{sec:our_method}}).
Table \ref{tab:metapaths} shows the three metapaths with the highest confidences.

\begin{table}
\caption{Three metapaths and their scores.}
\begin{center}
\resizebox{0.95\columnwidth}{!}{
\begin{tabular}{c|l}
$S(M)$ & Metapath $M$\\
\hline
\hline
0.446 & $(\textit{Compound} \xrightarrow{\textit{includes}^{-1}} \textit{Pharmacologic Class} \xrightarrow{\textit{includes}} \textit{Compound}$\\
& $\hspace{1mm} \xrightarrow{\textit{treats}} \textit{Disease})$\\
0.265 & $(\textit{Compound} \xrightarrow{\textit{resembles}} \textit{Compound} \xrightarrow{\textit{resembles}} \textit{Compound} \xrightarrow{\textit{treats}} \textit{Disease})$ \\
0.184 & $(\textit{Compound} \xrightarrow{\textit{binds}} \textit{Gene} \xrightarrow{\textit{associates}^{-1}} \textit{Disease})$
\end{tabular}
}
\label{tab:metapaths}
\end{center}
\end{table}

\if false
0.182 & $(\textit{Compound} \xrightarrow{\textit{resembles}} \textit{Compound} \xrightarrow{\textit{treats}} \textit{Disease})$\\
0.169 & $(\textit{Compound} \xrightarrow{\textit{palliates}} \textit{Disease} \xrightarrow{\textit{palliates}^{-1}} \textit{Compound} \xrightarrow{\textit{treats}} \textit{Disease})$\\
0.143 & $(\textit{Compound} \xrightarrow{\textit{binds}} \textit{Gene} \xrightarrow{\textit{binds}^{-1}} \textit{Compound} \xrightarrow{\textit{treats}} \textit{Disease})$\\
0.058 & $(\textit{Compound} \xrightarrow{\textit{causes}} \textit{Side Effect} \xrightarrow{\textit{causes}^{-1}} \textit{Compound} \xrightarrow{\textit{treats}} \textit{Disease})$\\
0.040 & $(\textit{Compound} \xrightarrow{\textit{treats}} \textit{Disease} \xrightarrow{\textit{resembles}} \textit{Disease})$\\
0.017 & $(\textit{Compound} \xrightarrow{\textit{resembles}} \textit{Compound} \xrightarrow{\textit{binds}} \textit{Gene} \xrightarrow{\textit{associates}^{-1}} \textit{Disease})$ \\
0.004 & $(\textit{Compound} \xrightarrow{\textit{binds}} \textit{Gene} \xrightarrow{\textit{expresses}^{-1}} \textit{Anatomy} \xrightarrow{\textit{localizes}^{-1}} \textit{Disease})$ \\
\fi

\subsection{Experimental Setup}
We apply our method, denoted by MINERVA+, to Hetionet and calculate hits@1, hits@3, hits@10, and the mean reciprocal rank (MRR).
During inference, a beam search is carried out, and the entities are ranked by the probability of their corresponding paths. 
Moreover, we consider another evaluation scheme (MINERVA+ (pruned)) that retrieves and ranks only those paths from the test rollouts that correspond to one of the metapaths.
All the other extracted paths are not considered in the ranking. 
We compare our approach with the path-based method MINERVA, the rule-based method AnyBURL, and the embedding-based methods TransE, RESCAL, and R-GCN.

\if false
For our method, the hyperparameter $\lambda$ that balances the two rewards (see Equation \ref{eq:rewards}) is chosen from the range $\{1,2,3,4,5\}$. The embedding size is tuned from $\{32, 50, 128\}$, and the hidden size of the MLP-layers for the action distribution is selected from $\{32, 50, 128, 256\}$. The number of LSTM-layers in the policy of the agent is chosen from $\{1, 2\}$, and the entropy regularizer $\beta$ takes values in $\{0.02, 0.05\}$. The path length is set to 3, the number of rollouts is given by 40 for training, and the beam width (number of test rollouts) at inference time is set to 100.

We compare our results with the following baseline methods. AnyBURL \cite{meilicke2019anytime} is a rule-based method that first mines logical rules by sampling paths in the graph and estimating the confidence of the rules. Then, it predicts candidates for link prediction, which are generated by applying the learned rules. 
We test AnyBURL also in an additional setting (AnyBURL (metapaths)): Instead of applying the learned rules for prediction, we use the rules derived from the metapaths in Table \ref{tab:metapaths} for inference.

The methods TransE \cite{bordes2013translating} and RESCAL \cite{nickel2011three} are both embedding-based models. TransE defines additive functions over the embeddings and employs distance-based scoring. RESCAL aims to compute a low-rank approximation of the adjacency tensor corresponding to a KG.
To cover a more recent paradigm in graph-based machine learning, we include a baseline graph convolution approach to compute KG embeddings given by the relational graph convolutional network R-GCN \cite{schlichtkrull2018modeling}. Extending the idea from the graph convolutional network GCN \cite{kipf2016semi} to multi-relational graphs, R-GCN mimicks the convolution operator on regular grids where an entity embedding is formed by aggregating node features from its neighbors. 



\fi

\subsection{Results}
\label{sec:results}

\begin{table}[h]
\caption{Comparison with baseline methods.
}
\begin{center}
\resizebox{0.95\columnwidth}{!}{
\begin{tabular}[H]{c c c c c}
Method & Hits@1 & Hits@3 & Hits@10 & MRR\\
\hline
\hline
AnyBURL & $0.139$ & $0.254$ & $0.358$ & $0.210$ \\
AnyBURL  (metapaths) & $0.252$ & $0.364$ & $0.609$ & $0.354$\\
TransE  & $0.073$ & $0.172$ & $0.318$ & $0.161$ \\
RESCAL  & $0.205$& $0.35$ &$0.576$ & $0.317$ \\
R-GCN  & $0.093$ &$0.245$ & $0.364$ & $0.188$ \\
MINERVA & $0.249$&$0.391$ & $0.605$& $0.357$\\
\hline
MINERVA+   & $0.294$& $0.437$&$0.615$ &$0.396$  \\
MINERVA+ (pruned)  & $\mathbf{0.319}$& $\mathbf{0.468}$&$\mathbf{0.628}$ &$\mathbf{0.416}$  \\
\end{tabular}
}
\label{tab:results}
\end{center}
\end{table}

Table \ref{tab:results} displays the test results for the experiments. 
The reported values for MINERVA and MINERVA+ correspond to the mean across five independent training runs. The standard errors lie between $0.0072$ and $0.0198$. This indicates that the reported performance gains are highly significant. 

AnyBURL only learns one rule for the relation $treats$ that has a length of at least 2. 
To see the effect of applying a larger number of rules, we try a setting where we use the metapaths for the prediction step, which leads to significantly improved results.
TransE and R-GCN show similar performance, and RESCAL performs best among the embedding-based methods.
Applying the modified ranking scheme, our method yields performance gains of $26.6\%$ for hits@1, $19.7\%$ for hits@3, $3.1\%$ for hits@10, and $16.5\%$ for MRR with respect to best performing baseline method.
 

\if false
\begin{figure}[t!]
    \centering
    \includegraphics[width=0.6\textwidth]{figures/rule_count_train_mean.png}
    \caption{Relative frequency of instance paths that correspond to a rule and lead to the correct entity during training. 
    }
    \label{fig:rule_count_train}
\end{figure}
\fi

\if false
The metapath that was most frequently extracted is 
\begin{align*}
    &(\textit{Compound} \xrightarrow{\textit{causes}} \textit{Side Effect} \xrightarrow{\textit{causes}^{-1}} \textit{Compound} \\ & \xrightarrow{\textit{treats}} \textit{Disease}) \;.
    \label{eq:CcSE_metapath}
\end{align*}
This rule was followed in $33.1\%$ of the paths during testing, of which $15.3\%$ ended at the correct entity. 
The second most extracted metapath is 
\begin{align*}
&(\textit{Compound} \xrightarrow{\textit{includes}^{-1}} \textit{Pharmacologic Class} \\ & \xrightarrow{\textit{includes}} \textit{Compound}\xrightarrow{\textit{treats}} \textit{Disease})
\end{align*}
with a confidence of $35.5\%$.
\fi


\if false
\begin{itemize}
    \item Add R-GCN \& metapath2vec
    \item Training efficiency (show training curves Minerva and 
    MINERVA+)
    \item Ablation study
\end{itemize}
\fi
\subsection{Discussion}
\label{subsec:discussion}
Our method can act as a generic mechanism to inject domain knowledge into reinforcement learning-based reasoning methods on KGs~\cite{lin2018multi,wenhan_emnlp2017}. While we employ rules that are extracted in a data-driven fashion, our method is agnostic towards the source of background information.
The additional reward for extracting a rule (see Equation~\eqref{eq:rewards}) can be considered as a regularization that enforces the agent to walk along metapaths that generalize to unseen instances. 

AnyBURL is strictly outperformed by both MINERVA and our method. Most likely, the large amount of high-degree nodes in Hetionet lead to the outcome that hardly any strong, predictive rules are extracted. 
Multi-hop reasoning methods contain a natural transparency mechanism by providing explicit inference  paths. 
Surprisingly, our experimental findings show that path-based reasoning methods outperform existing black-box methods on the drug repurposing task without a trade-off between explainability and performance. 
Both TransE and RESCAL are trained to minimize the reconstruction error in the immediate first-order neighborhood,
and our results indicate that these methods 
seem not to be suitable for the drug repurposing task.
R-GCN is in principle capable of modeling long-term dependencies due to the receptive field containing the entire set of nodes in the multi-hop neighborhood. However, the aggregation and combination step of R-GCN essentially acts as a low-pass filter on the incoming signals, and 
in the presence of many high-degree nodes, 
the center nodes may receive an uninformative signal that smooths over the neighborhood embeddings. 

To illustrate the applicability of our method, consider the compound sorafenib from Figure \ref{fig:kg_example}.
The three highest predictions of our model for new target diseases include hematologic cancer, breast cancer, and Barrett's esophagus. 
The database ClinicalTrails.gov \cite{national2007clinicaltrails} 
lists 23 clinical studies for testing the effect of sorafenib on these three diseases, showing that the predictions are meaningful targets for further investigation.

\if false
\begin{itemize}
    \item rule-based approaches and complexity
    \item policy-guided random walk vs nearest neighbor random walks
    \item R-GCN smoothes out important signals
    \item pretrained embeddings
    \item how important is the quality (confidence?) and quantity of the rules ablation study with a subset of the metapaths
    \item can we map our predictions to an existing medical study?
\end{itemize}
\fi

\section{Conclusion}
\label{sec:conclusion}
We have proposed a novel neuro-symbolic knowledge graph reasoning approach that leverages path-based reasoning,  representation learning, and logical rules. 
We apply our method to the highly relevant task of drug repurposing and compare our approach with both embedding-based and rule-based methods. We achieve better performance and an improvement of $26.6\%$ for hits@1 and $16.5\%$ for the mean reciprocal rank compared to popular baselines. 

\newpage
\section*{Acknowledgements}
This work has been supported by the German Federal Ministry for Economic Affairs and Energy (BMWi) as part of the project RAKI (no.\;01MD19012C).

\begin{figure}[h]
\centering
    \includegraphics[width=0.5\columnwidth]{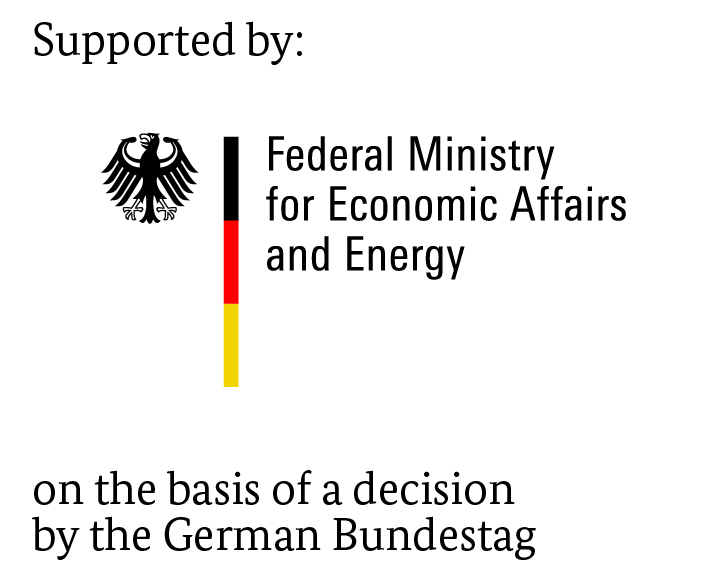}
\end{figure}

\bibliography{bibliography}

\begin{thebibliography}{15}
\providecommand{\natexlab}[1]{#1}
\providecommand{\url}[1]{\texttt{#1}}
\expandafter\ifx\csname urlstyle\endcsname\relax
  \providecommand{\doi}[1]{doi: #1}\else
  \providecommand{\doi}{doi: \begingroup \urlstyle{rm}\Url}\fi

\bibitem[Bordes et~al.(2013)Bordes, Usunier, Garcia-Dur\'{a}n, Weston, and
  Yakhnenko]{bordes2013translating}
Bordes, A., Usunier, N., Garcia-Dur\'{a}n, A., Weston, J., and Yakhnenko, O.
\newblock Translating embeddings for modeling multi-relational data.
\newblock In \emph{Proceedings of the 26th International Conference on Neural
  Information Processing Systems}, volume~2 of \emph{NIPS’13}, pp.\
  2787–2795, 2013.

\bibitem[Das et~al.(2018)Das, Dhuliawala, Zaheer, Vilnis, Durugkar,
  Krishnamurthy, Smola, and McCallum]{minerva}
Das, R., Dhuliawala, S., Zaheer, M., Vilnis, L., Durugkar, I., Krishnamurthy,
  A., Smola, A., and McCallum, A.
\newblock Go for a walk and arrive at the answer: reasoning over paths in
  knowledge bases using reinforcement learning.
\newblock In \emph{Proceedings of the 6th International Conference on Learing
  Representations}, 2018.

\bibitem[D{\"{o}}rpinghaus \& Jacobs(2019)D{\"{o}}rpinghaus and
  Jacobs]{Doerpinghaus2019}
D{\"{o}}rpinghaus, J. and Jacobs, M.
\newblock Semantic knowledge graph embeddings for biomedical research: data
  integration using linked open data.
\newblock In \emph{Proceedings of the Posters and Demo Track of the 15th
  International Conference on Semantic Systems (SEMANTiCS)}, volume 2451 of
  \emph{{CEUR} Workshop Proceedings}, 2019.

\bibitem[Hildebrandt et~al.(2020{\natexlab{a}})Hildebrandt, Li, Koner, Tresp,
  and Günnemann]{hildebr2020scene}
Hildebrandt, M., Li, H., Koner, R., Tresp, V., and Günnemann, S.
\newblock Scene graph reasoning for visual question answering.
\newblock \emph{arXiv:2007.01072}, 2020{\natexlab{a}}.

\bibitem[Hildebrandt et~al.(2020{\natexlab{b}})Hildebrandt, Serna, Ma,
  Ringsquandl, Joblin, and Tresp]{hildebrandt2020reasoning}
Hildebrandt, M., Serna, J. A.~Q., Ma, Y., Ringsquandl, M., Joblin, M., and
  Tresp, V.
\newblock Reasoning on knowledge graphs with debate dynamics.
\newblock In \emph{Proceedings of the 34th AAAI Conference on Artificial
  Intelligence}, 2020{\natexlab{b}}.

\bibitem[Himmelstein et~al.(2017)Himmelstein, Lizee, Hessler, Brueggeman, Chen,
  Hadley, Green, Khankhanian, and Baranzini]{himmelstein2017systematic}
Himmelstein, D.~S., Lizee, A., Hessler, C., Brueggeman, L., Chen, S.~L.,
  Hadley, D., Green, A., Khankhanian, P., and Baranzini, S.~E.
\newblock Systematic integration of biomedical knowledge prioritizes drugs for
  repurposing.
\newblock \emph{Elife}, 6:\penalty0 e26726, 2017.

\bibitem[Hochreiter \& Schmidhuber(1997)Hochreiter and
  Schmidhuber]{hochreiter1997long}
Hochreiter, S. and Schmidhuber, J.
\newblock Long short-term memory.
\newblock \emph{Neural Computation}, 9\penalty0 (8):\penalty0 1735--1780, 1997.

\bibitem[Lin et~al.(2018)Lin, Socher, and Xiong]{lin2018multi}
Lin, X.~V., Socher, R., and Xiong, C.
\newblock Multi-hop knowledge graph reasoning with reward shaping.
\newblock In \emph{Proceedings of the 2018 Conference on Empirical Methods in
  Natural Language Processing}, pp.\  3243--3253, 2018.

\bibitem[Meilicke et~al.(2018)Meilicke, Fink, Wang, Ruffinelli, Gemulla, and
  Stuckenschmidt]{meilicke2018fine}
Meilicke, C., Fink, M., Wang, Y., Ruffinelli, D., Gemulla, R., and
  Stuckenschmidt, H.
\newblock Fine-grained evaluation of rule- and embedding-based systems for
  knowledge graph completion.
\newblock In \emph{The Semantic Web -- ISWC 2018}, volume 11136 of
  \emph{Lecture Notes in Computer Science}, pp.\  3--20, 2018.

\bibitem[Meilicke et~al.(2019)Meilicke, Chekol, Ruffinelli, and
  Stuckenschmidt]{meilicke2019anytime}
Meilicke, C., Chekol, M.~W., Ruffinelli, D., and Stuckenschmidt, H.
\newblock Anytime bottom-up rule learning for knowledge graph completion.
\newblock In \emph{Proceedings of the 28th International Joint Conference on
  Artificial Intelligence}, pp.\  3137--3143, 2019.

\bibitem[Nickel et~al.(2011)Nickel, Tresp, and Kriegel]{nickel2011three}
Nickel, M., Tresp, V., and Kriegel, H.-P.
\newblock A three-way model for collective learning on multi-relational data.
\newblock In \emph{Proceedings of the 28th International Conference on Machine
  Learning}, 2011.

\bibitem[Schlichtkrull et~al.(2018)Schlichtkrull, Kipf, Bloem, Van Den~Berg,
  Titov, and Welling]{schlichtkrull2018modeling}
Schlichtkrull, M., Kipf, T.~N., Bloem, P., Van Den~Berg, R., Titov, I., and
  Welling, M.
\newblock Modeling relational data with graph convolutional networks.
\newblock In \emph{The Semantic Web -- ESWC 2018}, volume 10843 of
  \emph{Lecture Notes in Computer Science}, pp.\  593--607, 2018.

\bibitem[{U. S. National Library of
  Medicine}(2000)]{national2007clinicaltrails}
{U. S. National Library of Medicine}.
\newblock \url{clinicaltrails.gov}, 2000.

\bibitem[Williams(1992)]{williams1992simple}
Williams, R.~J.
\newblock Simple statistical gradient-following algorithms for connectionist
  reinforcement learning.
\newblock \emph{Machine Learning}, 8\penalty0 (3-4):\penalty0 229--256, 1992.

\bibitem[Xiong et~al.(2017)Xiong, Hoang, and Wang]{wenhan_emnlp2017}
Xiong, W., Hoang, T., and Wang, W.~Y.
\newblock {DeepPath}: A reinforcement learning method for knowledge graph
  reasoning.
\newblock In \emph{Proceedings of the 2017 Conference on Empirical Methods in
  Natural Language Processing}, pp.\  564--573, 2017.

\end{thebibliography}
\bibliographystyle{icml2020}

\end{document}